\documentclass[10pt,twocolumn,letterpaper]{article}

\usepackage{wacv}
\usepackage{times}
\usepackage{epsfig}
\usepackage{graphicx}
\usepackage{amsmath}
\usepackage{amssymb}
\usepackage{booktabs}
\usepackage{xcolor}
\usepackage{colortbl}

\def\wacvPaperID{RWS 14} 

\wacvalgorithmstrack   

\wacvfinalcopy 

\ifwacvfinal
\usepackage[breaklinks=true,bookmarks=false]{hyperref}
\else
\usepackage[pagebackref=true,breaklinks=true,colorlinks,bookmarks=false]{hyperref}
\fi
\usepackage[ruled,vlined,linesnumbered]{algorithm2e}

\begin{document}

\title{Iterative Scale-Up ExpansionIoU and Deep Features Association for Multi-Object Tracking in Sports}

\author{Hsiang-Wei Huang$^{1}$, Cheng-Yen Yang$^{1}$, Jiacheng Sun$^{1}$, Pyong-Kun Kim$^{2}$\\Kwang-Ju Kim$^{2}$, Kyoungoh Lee$^{2}$, Chung-I Huang$^{3}$, Jenq-Neng Hwang$^{1}$\\
$^{1}$Information Processing Lab, University of Washington\\
$^{2}$Electronics and Telecommunications Research Institute\\
$^{3}$National Center for High-Performance Computing\\
{\tt\small \{hwhuang,cycyang,sjc042,hwang\}@uw.edu}\\
{\tt\small \{iros,kwangju,longweek7\}@etri.re.kr}\\
\tt\small 1203033@narlabs.org.tw}

\maketitle
\thispagestyle{empty}

\begin{abstract}
Deep learning-based object detectors have driven notable progress in multi-object tracking algorithms. Yet, current tracking methods mainly focus on simple, regular motion patterns in pedestrians or vehicles. This leaves a gap in tracking algorithms for targets with nonlinear, irregular motion, like athletes. Additionally, relying on the Kalman filter in recent tracking algorithms falls short when object motion defies its linear assumption. To overcome these issues, we propose a novel online and robust multi-object tracking approach named deep ExpansionIoU (Deep-EIoU), which focuses on multi-object tracking for sports scenarios. Unlike conventional methods, we abandon the use of the Kalman filter and leverage the iterative scale-up ExpansionIoU and deep features for robust tracking in sports scenarios. This approach achieves superior tracking performance without adopting a more robust detector, all while keeping the tracking process in an online fashion. Our proposed method demonstrates remarkable effectiveness in tracking irregular motion objects, achieving a score of 77.2\% HOTA on the SportsMOT dataset and 85.4\% HOTA on the SoccerNet-Tracking dataset. It outperforms all previous state-of-the-art trackers on various large-scale multi-object tracking benchmarks, covering various kinds of sports scenarios. The code and models are available at \href{https://github.com/hsiangwei0903/Deep-EIoU}{https://github.com/hsiangwei0903/Deep-EIoU}.
\end{abstract}


\begin{figure}[t]
  \centering
  \includegraphics[width=\linewidth]{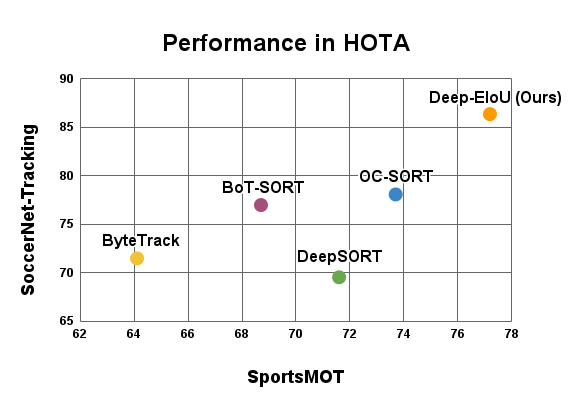}
  \caption{HOTA comparison of different trackers on the test sets of SoccerNet-Tracking and SportsMOT dataset. Deep-EIoU achieves 77.2\% HOTA on the SportsMOT test set and 85.4\% HOTA on the SoccerNet-Tracking test set. These results surpass the performance of all previous trackers on these large-scale multi-object tracking benchmarks. More comparisons between different trackers can be found in table \ref{table:sportsmot} and table \ref{table:soccernet}} 
  \label{fig:hota-two-dataset}
\end{figure}

\definecolor{aliceblue}{rgb}{0.94, 0.97, 1.0}


\section{Introduction}
Multi-Object Tracking (MOT) is a fundamental computer vision task that aims to track multiple objects in a video and localize them in each frame. Most recent tracking algorithms \cite{ByteTrack, aharon2022bot, DeepSORT, SORT}, which mainly focus on pedestrians or vehicle tracking, have achieved tremendous progress on public benchmarks \cite{mot16,dendorfer2020mot20,geiger2013vision}. However, these state-of-the-art algorithms fail to perform well on datasets with higher difficulties, especially those datasets with sports scenarios \cite{cui2023sportsmot,cioppa2022soccernet,zhao2023survey}. Given the growing demand for sports analytic for applications like automatic tactical analysis and athletes' movement statistics including running distance, and moving speed, the field of multi-object tracking for sports requires more attention.\\

\begin{figure}[t]
  \centering
  \includegraphics[width=\linewidth]{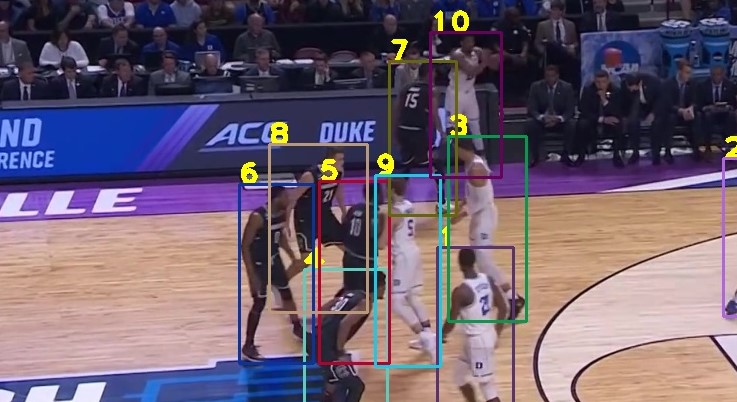}
  \caption{An example of the occlusion problem encountered during multi-athlete tracking. Occlusion can significantly hinder detection and tracking performance, and the occlusion issue in athlete tracking is particularly severe when compared to pedestrian tracking due to the high intensity of sports characteristics.}
  \label{fig:occlusion}
\end{figure}

\begin{figure}[t]
  \centering
  \includegraphics[width=1\linewidth]{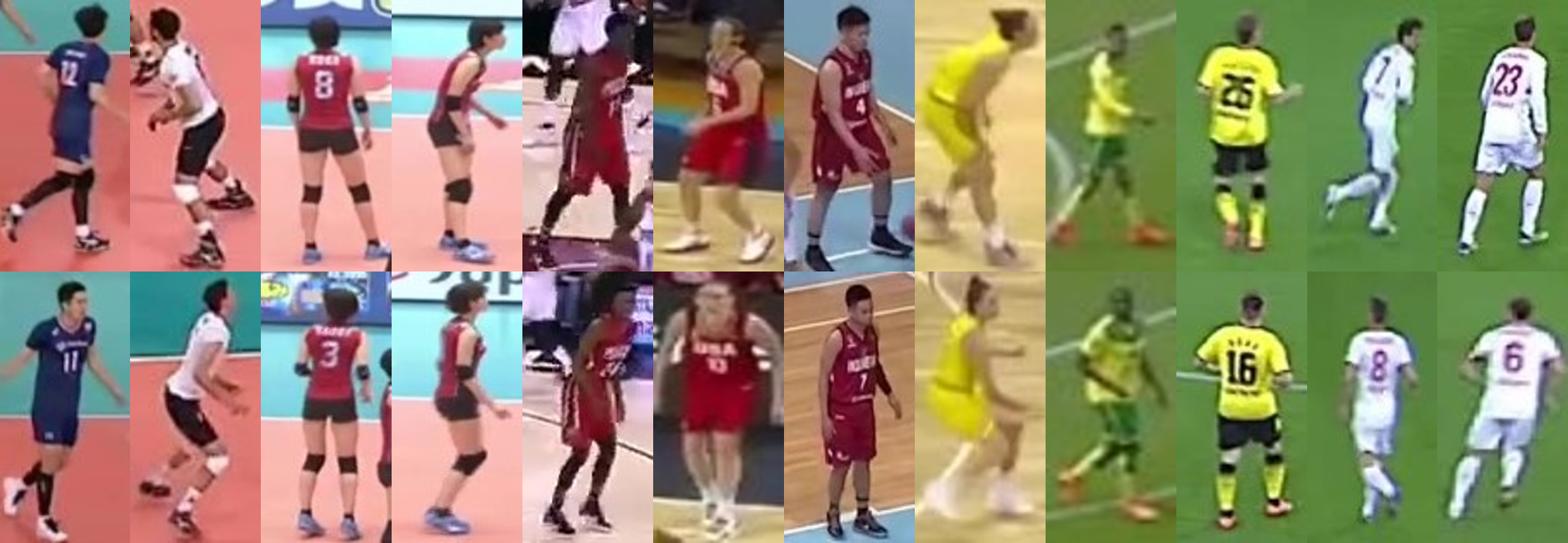}
  \caption{Example of similar appearances between the players from the SportsMOT dataset, which can cause confusion towards the tracker and decrease the tracking accuracy. Each column represents two different players with similar appearance.}
  \label{fig:similar}
\end{figure}

Different from multi-object tracking for pedestrians or vehicles, MOT in sports scenarios poses higher difficulties due to several reasons, including severe occlusion caused by the high intensity in sports scenes as illustrated in Figure \ref{fig:occlusion}, similar appearance between players in the same team due to the same color jersey like examples in Figure \ref{fig:similar}, and also unpredictable motion due to some sport movement like a crossover in basketball, sliding tackle in football or spike in volleyball. Due to the above reasons, the previous trackers, which utilize appearance-motion fusion \cite{FairMOT,DeepSORT} or simply motion-based \cite{ByteTrack,OCSORT,SORT} methods struggle to conduct robust tracking on several major MOT benchmarks in sports scenarios \cite{cioppa2022soccernet,cui2023sportsmot}.

To address these issues, in this paper, we propose a novel and robust online multi-object tracking algorithm specifically designed for objects with irregular and unpredictable motion. Our experimental results demonstrate that our algorithm effectively handles the irregular and unpredictable motion of athletes during the tracking process. It outperforms all tracking algorithms on two large-scale public benchmarks \cite{cui2023sportsmot} without introducing extra computational cost while maintaining the algorithm online. Therefore, in this paper, we assert three main contributions:

\begin{itemize}
\setlength{\itemsep}{0pt}
\setlength{\parsep}{0pt}
\setlength{\parskip}{0pt}
    \item We present a novel association method to specifically address the challenges in sports tracking, named ExpansionIoU, which is a simple yet effective method for tracking objects with irregular movement and similar appearances.
    \item Our proposed iterative scale-up ExpansionIoU further leverages with deep features association for robust multi-object tracking for sports scenarios. 
    \item The proposed method achieves $\textbf{77.2}$ HOTA on the SportsMOT \cite{cui2023sportsmot} dataset, and $\textbf{85.4}$ HOTA on the SoccerNet-Tracking dataset \cite{cioppa2022soccernet}, outperforming all the other previous tracking algorithms by a large margin.
\end{itemize}

\begin{figure*}[t]
  \centering
  \includegraphics[width=0.95\linewidth]{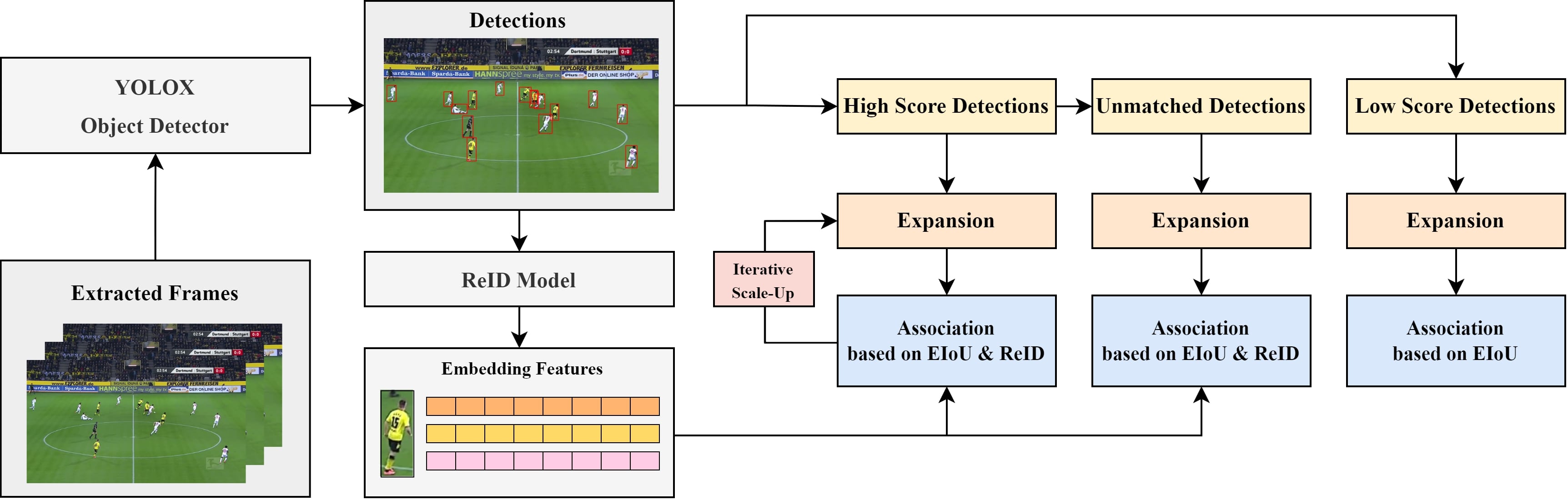}
  \caption{The proposed iterative scale-up ExpansionIoU tracking pipeline. The pseudo code of the proposed pipeline can be found in supplementary material.}
\end{figure*}

\section{Related Work}
\subsection{Multi-Object Tracking using Kalman Filter}
Most of the existing tracking algorithms \cite{ByteTrack,SORT,OCSORT,DeepSORT,zhang2021fairmot,yang2023multiobject,huang2023observation,huang2023multi,huang2023enhancing,yang2023sea} incorporate Kalman filter \cite{kf} as a method for object motion modeling. Kalman filter can formulate object motion as a linear dynamic system and can be used to predict its next frame location according to the object's motion from the previous frames. Kalman filter has shown effectiveness in multi-object tracking across several public benchmarks \cite{mot16,dendorfer2020mot20,sun2022dancetrack}. However, due to the Kalman filter's linear motion and Gaussian noise assumption, the Kalman filter might fail to track an object with nonlinear motion. Due to this reason, OC-SORT \cite{OCSORT} proposes several methods including observation-centric re-update to modify the Kalman filter's parameters during the tracking process and prevent error accumulations when an object is not tracked. The performance has shown effectiveness for tracking objects with irregular motion on several public datasets \cite{sun2022dancetrack,cui2023sportsmot}.

\subsection{Location-based Multi-Object Tracking}
Tracking can also be conducted based on the position information, given a high frame rate input video sequence, the object's position shift between frames is relatively small due to the high frame rates, thus making the position information a reliable clue for association between frames. Several methods \cite{stone2000centertrack,huang2023observation} utilizes the bounding boxes' distance as the cost for bounding box association, while some recent work \cite{yang2023hard} utilize different IoU calculation methods including GIoU \cite{rezatofighi2019generalized}, DIoU \cite{zheng2020distance}, and BIoU \cite{yang2023hard}, to conduct bounding box association between frames, which also demonstrate effectiveness in multi-object tracking.

\subsection{Appearance-based Multi-Object Tracking}
With the recent development and improvement of object ReID model \cite{OSNet} and training tricks \cite{luo2019bag}, many tracking algorithms incorporate ReID into the association process. Some methods use the joint detection and embedding architecture \cite{zhang2021fairmot,wang2020towards} to produce detection and object embedding at the same time to achieve real-time tracking. While the other methods \cite{DeepSORT,aharon2022bot} apply other stand-alone ReID model to extract detection's embedding features for association. The appearance-based tracking methods improve the tracking robustness with an extra appearance clue, while sometimes the appearance can be unreliable due to several reasons including occlusions, similar appearance among tracked objects, appearance variation caused by the object's rotation, or the lighting condition.

\subsection{Multi-Object Tracking in Sports}
Numerous studies have been conducted to monitor players' movements in team sports during games. This monitoring serves not only to automate the recording of game statistics but also enables sports analysts to obtain comprehensive information from a video scene understanding perspective. Different from MOT of pedestrian \cite{mot16}, MOT in sports scenarios is much more challenging due to several reasons including targets' faster and irregular motions, similar appearance among players in the same team, and more severe occlusion problem due to the sport's intense characteristic. The majority of recent methods for MOT for sports utilize the tracking-by-detection paradigm and integrate a re-identification network to generate an embedding feature for association.

Vats et al. \cite{hockey_ex_1} combine team classification and player identification approaches to improve the tracking performance in hockey. Similarly, Yang et al. \cite{soccer_ex_1} and Maglo et al. \cite{soccer_ex_3} demonstrate that by localizing the field and players, the tracking results in football can be more accurate. Additionally, Sang{\"{u}}esa et al. \cite{basketball_ex_1} utilize the human pose information and actions as the embedding features to enhance basketball player tracking. While Huang et al. \cite{huang2023observation} combine OC-SORT \cite{OCSORT} and appearance-based post-processing to conduct tracking on multiple sports scenarios including basketball, volleyball, and football \cite{cui2023sportsmot}.

\section{Proposed Methods}
Our proposed method follows the classic tracking-by-detection paradigm, which also enables online tracking without using future information. We first apply the object detector YOLOX on each input frame, and then we conduct association based on several clues including the similarity between extracted appearance features and the ExpansionIoU between the tracklets and detections. After the association cost is obtained, the Hungarian algorithm is conducted to get the best matching between tracklets and detections.

\subsection{Appearance-based Association}
The appearance similarity is a strong clue for object association between frames, the similarity can be calculated by the cosine similarity between the appearance features, and it can also be used to filter out some impossible associations. The cost for appearance association $Cost_A$ can be directly obtained from the cosine similarity with the following formula:
\begin{equation}
Cost_A = 1 - \text{Cosine Similarity} = 1 - \frac{a \cdot b}{\lVert a \rVert \lVert b \rVert}
\end{equation}

Here, $a$ and $b$ are the tracklet's appearance feature and the detection's appearance feature, respectively. A higher cosine similarity denotes a higher similarity in appearance, while a lower cosine similarity means the tracklet's appearance and the detection's appearance are different.

\subsection{Association with ExpansionIoU}
Insipired by previous work \cite{yang2023hard}, which utilizes expanding bounding boxes for association, to deal with the fast and irregular movement of sports player, we proposed ExpansionIoU (EIoU), a robust association method for tracking under large and nonlinear motion.
Different from the previous work \cite{yang2023hard}, we found out that expanding the bounding box even more during association can lead to a significantly better performance in athlete tracking.
Traditional IoU has been a cornerstone in location-based tracking method, but it often lacks the flexibility to account for object's large movement, when tracklet and detection bounding boxes share small or no IoU between adjacent frames. EIoU addresses this limitation by modifying the dimensions of bounding boxes, expanding their width and height and considers a wider range of object relationships, thus recover the association for those objects with large movement in sports scenarios. 
The expansion of bounding box is controlled by expansion scale $E$, given an original bounding box with height $h$ and width $w$, we can calculate the expansion length $h^{\star}$ and $w^{\star}$ following:

\begin{equation}
\begin{split}
h^{\star} &= (2E+1)h \\
w^{\star} &= (2E+1)w
\end{split}
\end{equation}

The original bounding box is expand based on the expansion length. Denote the original bounding box top-left and bottom-right coordinate as $(t,l)$,$(b,r)$, we can derive the expanded bounding box's coordinate as $(t-\frac{h^{\star}}{2},l-\frac{w^{\star}}{2})$ and $(b+\frac{h^{\star}}{2},r+\frac{w^{\star}}{2})$. 

The expanded bounding box is further used for IoU calculation between tracklets and detections pairs, note that the expansion is applied both on tracklets' last frame detections and the new coming detections from detector, the calculated EIoU is used for Hungarian association between adjacent frames. The operation of expanding the bounding box does not change several important objects' information like the bounding box center, aspect ratio, or appearance features. By simply expanding the search space, we can associate those tracklets and detections with small or no IoU, which is considered a common situation when the target's movement is fast, especially in sports games.

\subsection{Confidence Score Aware Matching}
Following ByteTrack \cite{ByteTrack}, we give the high confidence score detections higher weighting during the matching process. The high score detections usually imply less occlusion, hence a higher chance to preserve more reliable appearance features. Due to this reason, the first stage matching with high score detections is based on the association cost of both appearance and ExpansionIoU, denoted as $C_{stage1}$. The first stage of matching is built upon several rounds of iterative associations with a gradually scale-up expansion scale, addressed in Section \ref{sec:iterative}. In the second round of matching with low score detections, only ExpansionIoU is used, the cost is denoted as $C_{stage2}$. 

In our first matching stage, we abandon the IoU-ReID weighted cost method used in several previous works \cite{FairMOT,DeepSORT}, where the cost is a weighted sum of the appearance cost $C_A$ and IoU cost $C_{IoU}$:

\begin{equation}
C = \lambda C_A + (1-\lambda) C_{IoU}
\end{equation}

Instead, we adopt strategy similar to that of BoT-SORT \cite{aharon2022bot} for appearance-based association.  More specifically, we first filter out some impossible associations by setting cost thresholds for both appearance and ExpansionIoU (EIoU). The adjusted appearance cost $C_{\hat{A}}$ is set to 1 if either cost is bigger than its corresponding threshold, otherwise $C_{\hat{A}}$ is set as half of its appearance cost $C_A$. Finally, the first stage's final association cost $C_{stage1}$ is set as the minimum of the appearance cost $C_{\hat{A}}$ and EIoU cost $C_{EIoU}$. With $\tau_A$ and $\tau_{EIoU}$ denotes the threshold for the cost filter, we can write the appearance cost $C_{\hat{A}}$ as:

\begin{equation}
C_{\hat{A}} = \begin{cases}
1, & \text{if } C_{A} > \tau_A \text{ or } C_{EIoU} > \tau_{EIoU} \\
0.5 C_{A}, & \text{otherwise}
\end{cases}
\end{equation}

The final cost in the first stage of matching $C_{stage1}$ will be the minimum between adjusted appearance cost $C_{\hat{A}}$ and EIoU cost $C_{EIoU}$. 

\begin{equation}
C_{stage1} = \min(C_{\hat{A}}, C_{EIoU})
\end{equation}

While the association cost in the second matching stage $C_{stage2}$ will be only using the EIoU cost $C_{EIoU}$.

\subsection{Iterative Scale-Up ExpansionIoU}\label{sec:iterative}
As illustrated by the previous work using expansion bounding box for association \cite{yang2023hard}, the amount of the bounding box expansion is a crucial and sensitive hyperparameter in the tracking process and the performance of the tracker can be largely affected by the choice of the hyperparameter. In the real-world scenario, several factors might limit us from tuning the expansion scale and improving the tracking performance, including 1) the online tracking requirements. One common requirement for an athlete tracking system is the system needs to operate in an online matter, tuning the expansion scale with experiments and tweaking the performance is not possible in such cases. 2) No access to the testing data. For real-world scenarios, the testing data's ground truth is often not available, which makes finding the perfect expansion scale for association impossible. Due to the above reasons, we proposed a novel iterative scale-up ExpansionIoU association stage for robust tracking, the experiment results show that without any parameter tuning, our algorithms can always maintain SOTA performance on public benchmark.
Instead of doing hyperparameter tuning for the best expansion scale $E$, we choose to iteratively conduct EIoU association based on a gradually increasing $E_{t}$ during the tracking process. In each scale-up iteration, the expansion scale of the current iteration $E_{t}$ can be derived from the following formula:
\begin{equation}
    E_{t} = E_{initial} + \lambda t,
\end{equation}

\noindent where $E_{initial}$ is the initial expansion scale, $\lambda$ denotes the step size for the iterative scale-up process, $t$ stands for the iteration count, which starts from 0. By using this approach, we can first perform association to those trajectory and detection pairs with higher ExpansionIoU, and gradually search for those pairs with lower overlapping area, which enhances the robustness of our association process. Note that the iterative scale-up process is only applied for high score detections association, once the iteration count reaches the total number of iteration $t_{total}$, the association for high score detections stops and the tracker moves on to the low score detections association stage.

\section{Experiments and Results}

\subsection{Dataset}
We evaluate our tracking algorithm on two large-scale multi-sports player tracking datasets, i.e., SportsMOT \cite{cui2023sportsmot} and SoccerNet-Tracking \cite{cioppa2022soccernet}.\\
\begin{table}[h]
\small
\begin{tabular}{ccccc}
\hline
Sport Type & \# of tracks & \# of frames & Track Len & Density \\ \hline
Basketball & 10 & 845.4 & 767.9 & 9.1  \\
Football   & 22 & 673.9 & 422.1 & 12.8 \\
Volleyball & 12 & 360.4 & 335.9 & 11.2 \\ \hline
\end{tabular}
\caption{Summary of the SportsMOT dataset split by the type of sport. The number of tracks, number of frames, track length, and track density are average numbers across all videos of the sports. }
\end{table}

\noindent{\textbf{SportsMOT}} consists of 240 video sequences with over 150K frames and over 1.6M bounding boxes collected from 3 different sports, including basketball, football, and volleyball. Different from the MOT dataset \cite{mot16,dendorfer2020mot20}, SportsMOT possesses higher difficulties including: 1) targets' fast and irregular motions, 2) larger camera movements, and 3) similar appearance among players in the same team.\\
\noindent{\textbf{SoccerNet-Tracking}} is a large-scale dataset for multiple object tracking composed of 201 soccer game sequences. Each sequence is 30 seconds long. The dataset consists of 225,375 frames, 3,645,661 annotated bounding boxes, and 5,009 trajectories. Unlike SportsMOT, which only focuses on the tracking of sports players on the court, the tracking targets of SoccerNet contains multiple object classes including normal players, goalkeepers, referees, and soccer ball.

\subsection{Detector}
We choose YOLOX \cite{YOLOX} as our object detector to achieve real-time and high accuracy detection performance. Several existing trackers \cite{ByteTrack,OCSORT,aharon2022bot,yang2023hard} also incorporate YOLOX as detector, this also leads to a more fair comparison between these trackers with ours. We use the COCO pretrained YOLOX-X model provided by the official GitHub repositories of YOLOX \cite{YOLOX} and further fine-tune the model with SportsMOT training and validation set for 80 epochs, the input image size is 1440 $\times$ 800, with data augmentation including Mosaic and Mixup. We use SGD optimizer with weight decay of 5 $\times$ 10$-$4 and momentum of 0.9. The initial learning rate is 10$-$3 with 1 epoch warmup and cosine annealing schedule, which follows the same training procedure of ByteTrack's \cite{ByteTrack}. As for the SoccerNet-Tracking dataset, since oracle detections are provided in the dataset, to make a fair comparison and focus on tracking, we directly use the oracle detections provided by the dataset for the evaluation of all trackers.

\subsection{ReID Model}


For player re-identification (ReID), we use the omni-scale feature learning proposed in OSNet \cite{OSNet}. The unified aggregation gate fuses the features from different scales and enhances the ability of human ReID. \\
\noindent{\textbf{SportsMOT}} The ReID training data for experiments on SportsMOT dataset is constructed based on the original SportsMOT dataset where we crop out each player according to its ground truth annotation of the bounding boxes. The sampled dataset includes 31,279 training images, 133 query images, and 1,025 gallery images. \\
\noindent{\textbf{SoccerNet-Tracking}} We sample the ReID training data from the SoccerNet-Tracking training set, we randomly select 100 ground truth bounding boxes for each player from randomly sampled videos, with 65 used as training images, 10 used as query images, and 25 used as gallery images. The sampled ReID data contains 7,085 training images, 1,090 query images, and 2,725 gallery images, with a total of 109 randomly selected identities.\\
\noindent{\textbf{Training Details}} We use the pre-trained model from the Market-1501 dataset \cite{zheng2015scalable} and further fine-tune the model based on each of the above mentioned sampled sports ReID datasets, resulting in two ReID models for these two datasets. Each model is trained for 60 epochs, using Adam optimizer with cross entropy loss and the initial learning rate is 3 $\times$ 10$-$4. All the experiments are conducted on single Nvidia RTX 4080 GPU.\\

\subsection{Tracking Settings}
The threshold for detection to be treated as high score detection is 0.6, while detections with confidence score between 0.6 and 0.1 will be treated as low score detections, the rest detections with confidence score lower than 0.1 will be filtered. The cost filter threshold $\tau_A$ and $\tau_{EIoU}$ are set to 0.25 and 0.5, respectively. We also remove the constraint of aspect ratio in the detection bounding box, since sports scenarios might have the condition when a player is lying on the ground, which is different from the MOT datasets where most of the pedestrians are standing and walking. For the high score detections association, the initial value of expansion scale $E_{initial}$ is set to 0.7 with a step size $\lambda$ of 0.1, and the total number of iteration $t_{total}$ is 2. The expansion scale $E$ for low score detections association is 0.7, while for unmatched detections is 0.5. The max frames for keeping lost tracks is 60. After tracking is finished, linear interpolation is applied to boost the final tracking performance.

 



\begin{table*}[t]
  \begin{center}
    {\small{
\begin{tabular}{lccccccccc}
\toprule

 Method & Training Setup & HOTA$\uparrow$ & IDF1$\uparrow$ & AssA$\uparrow$ & MOTA$\uparrow$ & DetA$\uparrow$ & LocA$\uparrow$ & IDs$\downarrow$ & Frag$\downarrow$ \\
 
\midrule

FairMOT \cite{FairMOT} & Train & 49.3 & 53.5 & 34.7 & 86.4 & 70.2 & 83.9 & 9928 & 21673 \\
QDTrack \cite{fischer2022qdtrack} & Train & 60.4 & 62.3 & 47.2 & 90.1 & 77.5 & 88.0 & 6377 & 11850 \\
CenterTrack \cite{Zhou2020TrackingOA} & Train & 62.7 & 60.0 & 48.0 & 90.8 & 82.1 & 90.8 & 10481 & 5750 \\
TransTrack \cite{sun2020transtrack} & Train & 68.9 & 71.5 & 57.5 & 92.6 & 82.7 & 91.0 & 4992 & 9994 \\
BoT-SORT \cite{aharon2022bot} & Train & 68.7 & 70.0 & 55.9 & 94.5 & 84.4 & 90.5 & 6729 & 5349 \\
ByteTrack \cite{ByteTrack} & Train & 62.8 & 69.8 & 51.2 & 94.1 & 77.1 & 85.6 & 3267 & 4499 \\
OC-SORT \cite{OCSORT} & Train & 71.9 & 72.2 & 59.8 & 94.5 & 86.4 & 92.4 & 3093 & 3474 \\
ByteTrack \cite{ByteTrack} & Train+Val & 64.1 & 71.4 & 52.3 & 95.9 & 78.5 & 85.7 & 3089 & 4216 \\
OC-SORT \cite{OCSORT} & Train+Val & 73.7 & 74.0 & 61.5 & \textbf{96.5} & \textbf{88.5} & \textbf{92.7} & 2728 & 3144\\
MixSort-Byte \cite{cui2023sportsmot} & Train+Val & 65.7 & 74.1 & 54.8 & 96.2 & 78.8 & 85.7 & \textbf{2472} & 4009 \\
MixSort-OC \cite{cui2023sportsmot} & Train+Val & 74.1 & 74.4 & 62.0 & \textbf{96.5} & \textbf{88.5} & 92.7 & 2781 & 3199\\
\midrule
\rowcolor{aliceblue}
Deep-EIoU (Ours) & Train & 74.1 & 75.0 & 63.1 & 95.1 & 87.2 & 92.5 & 3066 & 3471 \\
\rowcolor{aliceblue}
Deep-EIoU (Ours) & Train+Val & \textbf{77.2} & \textbf{79.8} & \textbf{67.7} & 96.3 & 88.2 & 92.4 & 2659 & \textbf{3081} \\

\bottomrule
\end{tabular}
}}
\end{center}
\caption{The performance comparison between different state-of-the-art trackers on the SportsMOT test sets. Our algorithm outperforms all the other previous tracking algorithms and achieves SOTA performance in several major evaluation metrics including HOTA, IDF1, and AssA. The evaluation results besides BoT-SORT are taken from the number reported in the SportsMOT dataset paper \cite{cui2023sportsmot}. While BoT-SORT is evaluated based on their official code \cite{aharon2022bot}.
} 
\label{table:sportsmot}
\end{table*}

\subsection{Evaluation Metrics}
MOTA \cite{mota} is often used as an evaluation metric for multi-object tracking task, however, MOTA mainly focuses on the detection performance instead of association accuracy. Recently, in order to balance between the detection and association performance, more and more public benchmarks start to use HOTA \cite{luiten2021hota} as the main evaluation metric. For evaluation on the SportsMOT dataset, we adopt HOTA, MOTA, IDF1, and other associated metrics \cite{bernardin2008evaluating} for comparison. While for SoccerNet, we adopt HOTA metrics, with associated DetA, and AssA metrics, since only these metrics are provided by the evaluation server.

\subsection{Performance}
We compare our tracking algorithm with previous existing trackers on two large-scale multi-object tracking datasets in sports scenarios, the SportsMOT and SoccerNet-Tracking datasets. All the experiments are run on one Nvidia RTX 4080 GPU, and the tracking results are evaluated on the datasets' official evaluation server.

\noindent{\textbf{SportsMOT}} As shown in table \ref{table:sportsmot}, the performance of our proposed Deep-EIoU achieves $\textbf{77.2}$ in HOTA, $\textbf{79.8}$ in IDF1, $\textbf{67.7}$ in AssA. The performance of our method achieves state-of-the-art results and outperforms all the other previous trackers while also keeping the tracking process in an online fashion, showing the effectiveness of our algorithm in multi-object tracking in sports scenarios.\\

\noindent{\textbf{SoccerNet}} To focus on the tracking performance and make a fair comparison, all the evaluated methods are using oracle detections provided by the SoccerNet-Tracking dataset \cite{cioppa2022soccernet}. The performance of our proposed method is reported in table \ref{table:soccernet}. Our method achieves $\textbf{85.443}$ in HOTA, $\textbf{73.567}$ in AssA, $\textbf{99.236}$ in DetA, which outperforms several state-of-the-art online tracking algorithms by a large margin. The performance of DeepSORT and ByteTrack are reported from the original SoccerNet-Tracking paper \cite{cioppa2022soccernet}. The competitive performance of Deep-EIoU in various large-scale sports player tracking datasets demonstrates the effectiveness of our algorithm in multi-object tracking in sports.

\begin{table}[h]
\centering
\begin{tabular}{lccc}
\hline
Tracker & HOTA & AssA & DetA \\ \hline
DeepSORT \cite{DeepSORT}          & 69.552 & 58.668 & 82.628  \\
ByteTrack \cite{ByteTrack}        & 71.500 & 60.718 & 84.342  \\
BoT-SORT \cite{aharon2022bot}     & 76.999 & 63.447 & 93.525  \\
OC-SORT  \cite{OCSORT}             & 78.091 & 64.687 & 94.273  \\
\rowcolor{aliceblue}
Deep-EIoU (Ours)  & \textbf{85.443} & \textbf{73.567} & \textbf{99.236} \\
\hline
\end{tabular}
\caption{Performance comparison of different tracking methods using oracle detections on the SoccerNet-Tracking \cite{cioppa2022soccernet} test set. The performance of DeepSORT and ByteTrack are reported from the SoccerNet-Tracking dataset paper \cite{cioppa2022soccernet}. While BoT-SORT and OC-SORT are evaluated using their official code.}
\label{table:soccernet}
\end{table}

\begin{table}[h]
\centering
\begin{tabular}{ccccc}
\hline
Method   & ReID         & ISU          & LI           & HOTA ($\uparrow$) \\ \hline
Baseline & -            & -            & -            & 71.403\\
-        & $\checkmark$ &              &              & 75.266\\
-        & $\checkmark$ & $\checkmark$ &              & 77.205\\
-        & $\checkmark$ & $\checkmark$ & $\checkmark$ & 77.220\\
\hline
\end{tabular}
\caption{We evaluate the Deep-EIoU baseline with different settings on the SportsMOT test set. Including using the ReID model for association based on appearance, Iterative Scale-Up (ISU) process and using Linear Interpolation (LI) as post-processing for our method.}
\label{table:add}
\end{table}


\subsection{Ablation Studies on Deep-EIoU}
In our experiments, Deep-EIoU is evaluated with different settings on the SportsMOT test set, including whether to incorporate appearance (ReID) during tracking, using iterative scale-up bounding box expansion, and using linear interpolation as post-processing.  As shown in Table \ref{table:add}, after incorporating ReID model based on appearance association, the HOTA of Deep-EIoU is boosted by 3.8, showing that although sharing similar appearance between athletes, it is still important to use appearance as a clue for tracking in sport scenarios. With the iterative scale-up process (ISU), the gradually scale-up bounding box can first establish association with those tracklets and detections with higher EIoU, thus also increase the tracking performance, note that the iterative scale-up process is incorporate with a larger tracking buffer, unlike the default setting of 30 for pedestrian tracking, we use 60 due to the stronger occlusion characteristics of the sports scenarios. And finally, following most of the online tracking algorithm \cite{ByteTrack,OCSORT}, we also include linear interpolation (LI) as a strategy to boost the final tracking performance.

\begin{figure}[h]
\includegraphics[width=\linewidth]{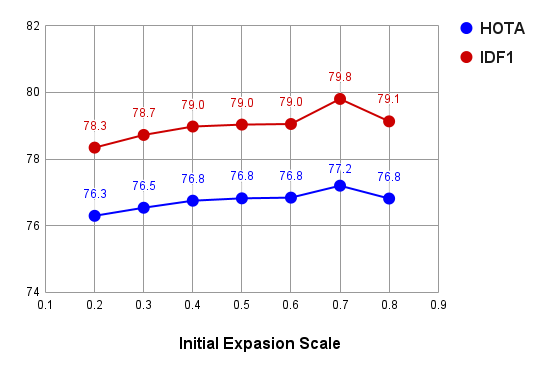}
\centering
\caption{Performance comparison of Deep-EIoU under different initial expansion scales on the SportsMOT test set.}
\label{fig:initial}
\end{figure}

\subsection{Robustness to initial expansion scale}
To prove the effectiveness and robustness of our approach, we conduct experiments based on different initial expansion scales in the iterative scale-up process. We change the initial expansion scale from 0.2 to 0.8. The experiment results in Figure \ref{fig:initial} show that we can still achieve SOTA performance with different initial expansion scales because the iterative scale-up process can enhance the robustness and does not require any parameter tuning to achieve SOTA performance. This proves our method's effectiveness in the real-world scenario, when ground truth is often not available and the tracking parameter can not be tuned.

\begin{table}[h]
\centering
\begin{tabular}{lcccc}
\hline
Tracker   & w/ EIoU & HOTA & AssA & DetA \\ \hline
ByteTrack &              & 62.8 & 51.2 & 77.1\\
ByteTrack & $\checkmark$ & 67.5 & 54.4 & 83.9\\
BoT-SORT   &              & 68.7 & 55.9 & 84.4\\
BoT-SORT   & $\checkmark$ & 71.3 & 60.2 & 84.5\\
\hline
\end{tabular}
\caption{We evaluate two classic Kalman filter-based tracking algorithms including ByteTrack \cite{ByteTrack} and BoTSORT \cite{aharon2022bot} on the SportsMOT test set. Experiment results show that the Kalman filter-based tracker can also be benefited from incorporating ExpansionIoU during the tracking process.}
\label{table:kf-eiou}
\end{table}

\subsection{ExpansionIoU on Kalman filter-based tracker}

To test the effect of ExpansionIoU on the Kalman filter-based tracker, we also implement several versions of our method by directly incorporating the Kalman filter and ExpansionIoU. In our implementation, the Kalman filter's prediction and detection will be expanded in the tracking process following the ExpansionIoU. The experiment results in Table \ref{table:kf-eiou} demonstrate that after directly replacing IoU with EIoU, these two classic Kalman filter-based trackers increase their performance by a large margin in HOTA, AssA, and DetA. This demonstrates that ExpansionIoU can also be applied as a plug-and-play trick for Kalman filter-based tracker to boost the tracking performance.

\begin{figure*}[t]
\includegraphics[width=\textwidth]{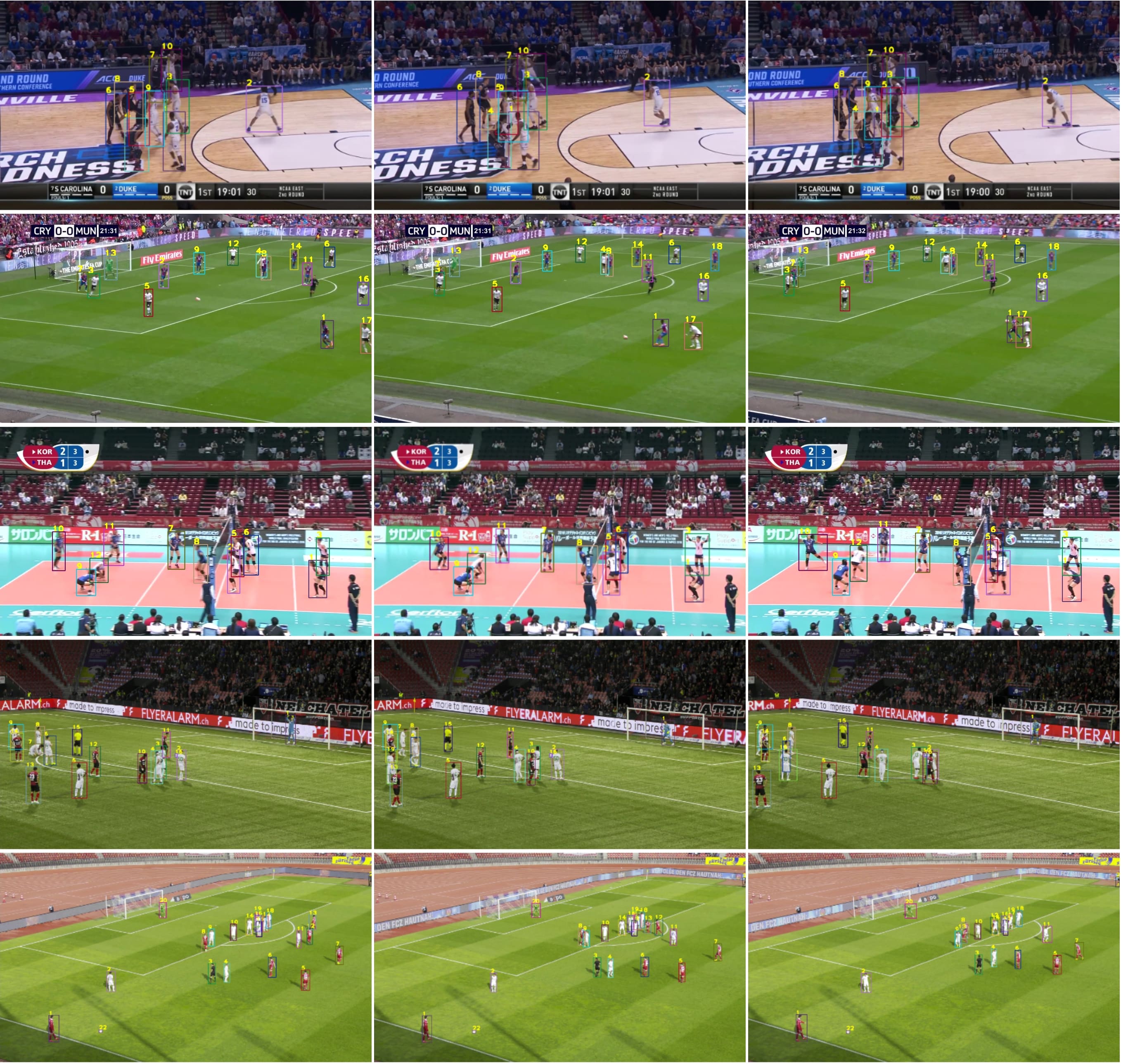}
\centering
\caption{Visualization results of Deep-EIoU from random sampled clips of SportsMOT dataset (row 1 to 3) and SoccerNet-Tracking dataset (row 4 to 5). With the iterative scale-up ExpansionIoU and deep features association, our algorithm can achieve robust multi-athlete tracking under severe occlusion conditions in multiple diverse sports scenarios including basketball, football, and volleyball. More visualization results can be found in supplementary material.}
\end{figure*}


\subsection{Limitations}
While our algorithm provides a robust and practical solution for online multi-object tracking in sports scenarios, it does have its limitations, including the absence of an offline post-processing trajectories refinement method. Such methods could involve a post-processing approach \cite{huang2023observation} or a strong memory buffer \cite{wang2022sportstrack}, which would be valuable in handling edge cases where sports players temporarily exit and re-enter the camera's field of view. It is worth noting that exploring and integrating offline refinement techniques in the future could potentially enhance the overall performance and extend the applicability of our approach beyond short-term tracking scenarios.

Another concern of Deep-EIoU is its relatively slower running speed when compared with motion-based trackers. Despite delivering significantly enhanced performance, the integration of the appearance-based tracking-by-detection framework, which involves a detector and a ReID model, introduces additional computational cost. The current Deep-EIoU pipeline achieves around 14.6 FPS on a single Nvidia RTX 4080 GPU, which is slower compared to motion-based method. It's worth noting that transitioning to a more lightweight detector and ReID model has the potential to significantly boost operational speed.

\section{Conclusions}
In this paper, we proposed Deep-EIoU, an iterative scale-up ExpansionIoU and deep features association method for multi-object tracking in sports scenarios, which achieves competitive performance on two large-scale multi-object sports player tracking datasets including SportsMOT and SoccerNet-Tracking. Our method successfully tackles the challenges of irregular movement during multi-object tracking in sports scenarios and outperforms the previous tracking algorithms by a large margin.\\

{\small
\bibliographystyle{ieee_fullname}
\bibliography{egbib}
}

\end{document}